\def\year{2016}\relax
\documentclass[letterpaper]{article}
\usepackage{aaai16}
\usepackage{times}
\usepackage{helvet}
\usepackage{courier}

\usepackage{amsmath}
\usepackage{graphicx}

\begin{document}
%
\title{Automatic Detection and Categorization of Election-Related Tweets}
\author{Prashanth Vijayaraghavan\thanks{The first two authors contributed equally to this work.}, Soroush Vosoughi\footnotemark[1], Deb Roy \\ Massachusetts Institute of Technology \\ Cambridge, MA, USA \\ \tt{pralav@mit.edu, soroush@mit.edu, dkroy@media.mit.edu} }
\maketitle
\begin{abstract}
\begin{quote}
With the rise in popularity of public social media and micro-blogging services, most notably Twitter, the people have found a venue to hear and be heard by their peers without an intermediary. As a consequence, and aided by the public nature of Twitter, political scientists now potentially have the means to analyse and understand the narratives that organically form, spread and decline among the public in a political campaign.
However, the volume and diversity of the conversation on Twitter, combined with its noisy and idiosyncratic nature, make this a hard task. Thus, advanced data mining and language processing techniques are required to process and analyse the data. In this paper, we present and evaluate a technical framework, based on recent advances in deep neural networks, for identifying and analysing election-related conversation on Twitter on a continuous, longitudinal basis. Our models can detect election-related tweets with an F-score of 0.92 and can categorize these tweets into 22 topics with an F-score of 0.90. 
\end{quote}
\end{abstract}

\linespread{.97}
\setlength{\belowcaptionskip}{-6pt}

\section{Introduction}
With the advent and rise in popularity of public social media sites--most notably Twitter--the people have a venue to directly participate in the conversation around the elections. Recent studies have shown that people are readily taking advantage of this opportunity and are using Twitter as a political outlet, to talk about the issues that they care about in an election cycle \cite{twitter_elec}.

The public nature, and the sheer number of active users on Twitter, make it ideal for \emph{in vivo} observations of what the public cares about in the context of an election (as opposed to \emph{in vitro} studies done by polling agencies). To do that, we need to trace the election’s prevailing narratives as they form, spread, morph and decline among the Twitter public. Due to the sheer volume of the Twitter conversation around the US presidential elections, both on Twitter and traditional media, advanced data mining and language processing techniques are required to ingest, process and analyse the data. In this paper, we present and evaluate a technical framework, based on recent advances in deep neural networks, for identifying and analysing election-related conversation on Twitter on a continuous, longitudinal basis. The framework has been implemented and deployed for the 2016 US presidential election, starting from February 2015.

Our framework consists of a processing pipelines for Twitter. Tweets are ingested on a daily basis and passed through an election classifier. Next, tweets that have been identified as election-related are passed to topic and sentiment classifiers to be categorized. The rest of this paper explains each of these steps in detail. 





\section{Detecting Election-Related Tweets}
According to Twitter estimates, there are more than half a billion tweets sent out daily. In order to filter for election-related tweets, we first created a ``high precision" list of 86 election-related seed terms. These seed terms included election specific terms (e.g., \#election2016, 2016ers, etc) and the names and Twitter handles of the candidates (eg. Hillary Clinton, Donald Trump, @realDonaldTrump, etc). These are terms that capture election-related tweets with very few false positives, though the recall of these terms is low. 

In order to increase the recall further, and to capture new terms as they enter the conversation around the election, on a weekly basis we expand the list of terms using a query expansion technique. This increases the number of tweets extracted by about 200\%, on average. As we show later in the paper, this corresponds to an increase in the recall (capturing more election-related tweets), an a decrease in the precision. Finally, in order to increase the precision, the tweets extracted using the expanded query are sent through a tweet election classifier that makes the final decision as to whether a tweet is about the election or not. This reduces the number of tweets from the last stage by about 41\%, on average. In the following sections we describe these methods in greater detail. 

\subsection{Query Expansion}
On a weekly basis, we use the Twitter historical API (recall that we have access to the full archive) to capture all English-language tweets that contain one or more of our predefined seed terms. For the query expansion, we employ a continuous distributed vector representation of words using the continuous Skip-gram model (also known as $Word2Vec$) introduced by Mikolov et al. \cite{mikolov2013distributed}. The model is trained on the tweets containing the predefined seed terms to capture the context surrounding each term in the tweets. This is accomplished by maximizing the objective function:
\begin{equation}\label{wordvec}\dfrac{1}{|V|}\sum_{n=1}^{|V|}\sum_{-c\leq j\leq c,j\neq 0}{\log  p(w_{n+j}|w_n)}\end{equation}
where $|V| $ is the size of the vocabulary in the training set and $c$ is the size of context window. The probability $p(w_{n+j}|w_n)$ is approximated using the hierarchical softmax introduced and evaluated by Morin and Bengio \shortcite{morin2005hierarchical}. The resultant vector representation captures the semantic and syntactic information of the all the terms in the tweets which, in turn, can be used to calculate similarity between terms. 



Given the vector representations for the terms, we calculate the similarity scores between pairs of terms in our vocabulary using cosine similarity. For a term to be short-listed as a possible extension to our query, it needs to be mutually similar to one of the seed terms (i.e., the term needs to be in the top 10 similar terms of a seed term and vice versa). The top 10 mutually similar terms along with noun phrases containing those terms are added to the set of probable election query terms. 

However, there could potentially be voluminous conversations around each of the expanded terms with only a small subset of the conversation being significant in the context of the election. Therefore, the expanded terms are further refined by extracting millions of tweets containing the expanded terms, and measuring their election significance metric $\rho$. This is the ratio of number of election-related tweets (measured based on the presence of the seed terms in the tweets) to the total number of retrieved tweets. For our system, the terms with $\rho \geq 0.3$ form the final set of expanded query terms. The $\rho$ cut-off can be set based on the need for precision or recall. We set the cut-off to be relatively low because as we discuss in the next section, we have another level of filtering (an election classifier) that further increases the precision of our system. 

The final list of terms generated usually includes many diverse terms, such as terms related to the candidates, their various Twitter handles, election specific terms (e.g., \#election2016), names and handles of influential people and prominent issue-related terms and noun phrases (e.g., immigration plan). As mentioned earlier, the query is expanded automatically on a weekly basis to capture all the new terms that enter the conversation around the election (for example the hashtag, \#chaospresident , was introduced to the Twitter conversation after the December 15, 2015 republican debate). Table \ref{tab:expand_1} shows a few examples of expanded terms for some of the candidates.


\begin{table}[h]
\centering
\small
\begin{tabular}{|l|l|}
\hline %
\emph{Candidate Seed Term} & \emph{Exapanded Terms} \\\hline 
Mike Huckabee & ckabee, fuckabee, hucka, wannabee \\\hline
Carly Fiorina & failorina, \#fiorinas, dhimmi,fiorin\\\hline
Hillary Clinton &  \#hillaryforpotus, hitlery, hellary \\\hline
Donald Trump &  \#trumptheloser, \#trumpthefascist \\\hline
\end{tabular}
\caption{Examples of expanded terms for a few candidates.}
\label{tab:expand_1}
\end{table}

\subsection{Election Classifier}
The tweets captured using the expanded query method include a number of spam and non-election-related tweets. In most cases, these tweets contain election-related hashtags or terms that have been maliciously put in non-election related tweets in order to increase their viewership. The election classifier acts as a content-aware filter that removes non-election and spam tweets from the data captured by the expanded query. 

Because of the noisy and unstructured nature of tweets, we use a deep character-level election classifier. Character-level models are great for noisy and unstructured text since they are robust to errors and misspellings in the text. Our classifier models tweets from character level input and automatically learns their abstract textual concepts. For example, our character-level classifier would closely associate the words ``no'' and ``noooo'' (both common on twitter), while a word-level model would have difficulties relating the two words. 

\begin{figure}[]
\includegraphics[width=\columnwidth]{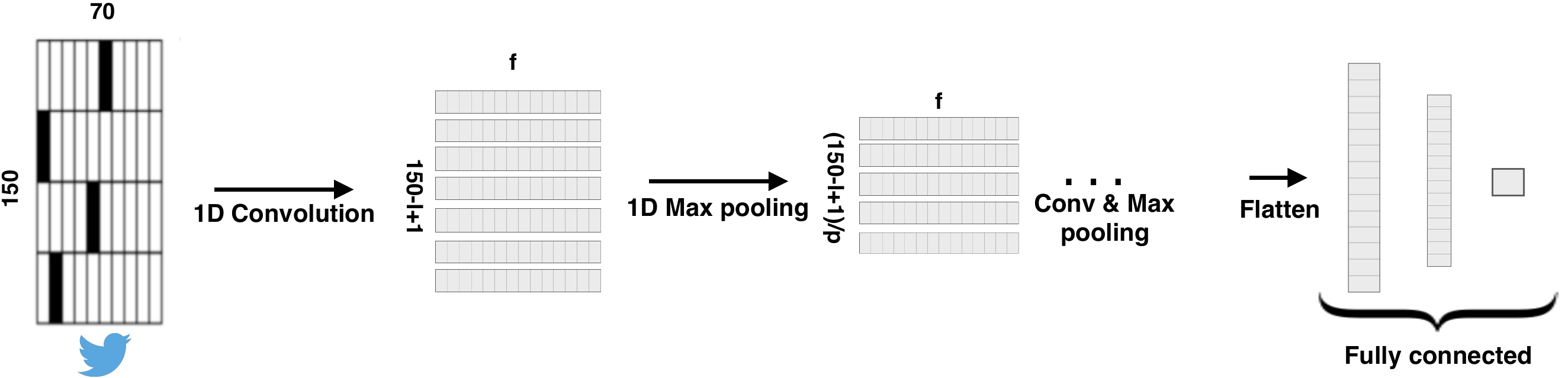} 
\caption{Tweet Election Model -  Deep Character-level Convolutional Neural Network (150 is maximum number of characters in a tweet, 70 is the character set size, $l$ is the character window size,  $f$ is the number of filters, $p$ is the pooling size).}
\label{cnn_elect}
\end{figure}

The model architecture can be seen in Figure \ref{cnn_elect}. This is a slight variant of the deep character level convolutional neural network introduced by Zhang et al \cite{zhang2015text}. We adapted their model to work with short text with a predefined number of characters, such as tweets with their 140 character limit. The character set considered for our classification includes the English alphabets, numbers, special characters and unknown character (70 characters in total). 


Each character in a tweet is encoded using a one-hot vector $x_i\in\{0,1\}^{70}$. Hence, each tweet is represented as a binary matrix $x_{1..150}\in\{0,1\}^{150\times70}$ with padding wherever necessary, where 150 is the maximum number of characters in a tweet (140 plus 10 for padding) and 70 is the size of the character set shown above. 
Each tweet, in the form of a 150$\times$70 matrix, is fed into a deep model consisting of five 1-d convolutional layers and three fully connected layers. A convolution operation employs a filter $w$, to extract l-gram character feature from a sliding window of $l$ characters at the first layer and learns abstract textual features in the subsequent layers. This filter $w$ is applied across all possible windows of size $l$ to produce a feature map. A sufficient number ($f$) of such filters are used to model the rich structures in the composition of characters. Generally, with tweet s, each element $c_i^{(h,F)}(s)$ of a feature map $F$ at the layer $h$ is generated by:

\begin{equation}\label{conv}c_i^{(h,F)}(s)=g(w^{(h,F)}\odot \hat{c}_i^{(h-1)}(s)+b^{(h,F)}) 
\end{equation}

where 
$w^{(h,F)}$ is the filter associated with feature map F at layer $h$; 
$\hat{c}_i^{(h-1)}$ denotes the segment of output of layer $h-1$ for convolution at location $i$; $b^{(h,F)}$ is the bias associated with that filter at layer $h$; $g$ is a rectified linear unit and $\odot$ is element-wise multiplication. The output of the convolutional layer $c^h(s)$ is a matrix, the columns of which are feature maps $c^{(h,F_k)}(s)  \vert  k \in 1..f$. The output of the convolutional layer is followed by a 1-d max-overtime pooling operation \cite{collobert2011natural} over the feature map and selects the maximum value as the prominent feature from the current filter. Pooling size may vary at each layer (given by $p^{(h)}$ at layer $h$). The pooling operation shrinks the size of the feature representation and filters out trivial features like unnecessary combination of characters (in the initial layer). The window length $l$, number of filters $f$, pooling size $p$ at each layer are given in Table \ref{layers}.


\begin{table}[h]
\centering
\small
\begin{tabular}{ |c|c|c|c|c| }
\hline %
 \emph{Layer} & \emph{Input}  & \emph{Window}  & \emph{Filters} & \emph{Pool}  \\
\emph{(h)}& \emph{Size} & \emph{Size} \emph{(l)} & \emph{(f)}& \emph{Size} \emph{(p)} \\\hline %
$1$ & $150\times 70$ & $7$ & $256$ & $3$ \\\hline
$2$ &$42\times 256$ &7 &$256$ & $3$ \\\hline
$3$& $14\times 256$ &  $3$ & $256$ & $N/A$\\\hline
$4$ & $12\times 256$ &$3$ & $256$ & $N/A$ \\\hline
$5$& $10\times 256$ &$3$ & $256$ & $N/A$ \\\hline
\end{tabular}
\caption{Convolutional layers with non-overlapping pooling layers used for the election classifier.}
\label{layers} %
\end{table}

The output from the last convolutional layer is flattened. The input to the first fully connected layer is of size 2048 ($8\times 256$). This is further reduced to vectors of sizes 1024 and 512 with a single output unit, where the sigmoid function is applied (since this is a binary classification problem). For regularization, we applied a dropout mechanism after the first fully connected layer. This prevents co-adaptation of hidden units by randomly setting a proportion $\rho$ of the hidden units to zero (for our case, we set $\rho=0.5$). The objective was set to be the binary cross-entropy loss (shown below as $BCE$):
\begin{equation}BCE(t,o)=-t\log(o) -\\ (1-t)\log(1-o)\end{equation}
where $t$ is the target and $o$ is the predicted output. The \emph{Adam Optimization} algorithm \cite{kingma2014adam} is used for learning the parameters of our model. 

The model was trained and tested on a dataset containing roughly 1 million election-related tweets and 1 million non-election related tweets. These tweets were collected using distant supervision. The ``high precision'' seeds terms explained in the previous section were used to collect the 1 million election-related tweets and an inverse of the terms was used to collect 1 million non-election-related tweets. The noise in the dataset from the imperfect collection method is offset by the sheer number of examples. Ten percent of the dataset was set aside as a test set. The model performed with an F-score of $0.99$ ($.99$ precision, $.99$ recall) on the test data.

\subsection{Evaluation} We evaluated the full Twitter ingest engine on a balanced dataset of 1,000 manually annotated tweets. In order to reduce potential bias, the tweets were selected and labelled by an annotator who was familiar with the US political landscape and the upcoming Presidential election but did not have any knowledge of our system. The full ingest engine had an F-score of $0.92$, with the precision and recall for the election-related tweets being $0.91$ and $0.94$ respectively.

Note that the evaluation of the election classifier reported in the last section is different since it was on a dataset that was collected using the election related seed terms, while this evaluation was done on tweets manually selected and annotated by an unbiased annotator. 

\section{Topic-Sentiment Convolutional Model}
The next stage of our Twitter pipeline involves topic and sentiment classification. With the help of an annotator with political expertise, we identified 22 election-related topics that capture the majority of the issue-based conversation around the election. These topics are: Income Inequality, Environment/Energy, Jobs/Employment, Guns, Racial Issues, Foreign Policy/National Security, LGBT Issues, Ethics, Education, Financial Regulation, Budget/Taxation, Veterans, Campaign Finance, Surveillance/Privacy, Drugs, Justice, Abortion, Immigration, Trade, Health Care, Economy, and Other.

\begin{figure}[]
\centering
\includegraphics[width=\columnwidth]{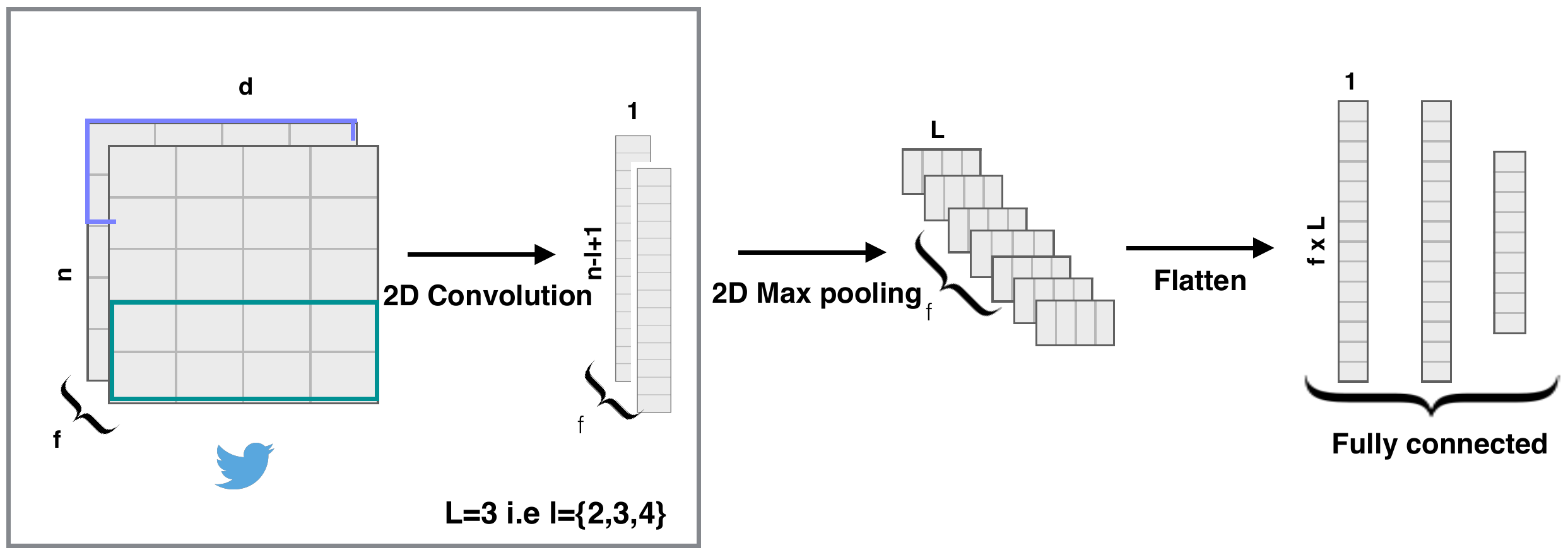}
\caption{Tweet topic-sentiment convolutional model (n=50 is maximum number of words in a tweet, d is the word embedding dimension, $L$ is the number of n-grams, $f$ is the number of filters, $(n-ngram+1\times 1)$ is the pooling size).}
\label{topic_cnn}
\end{figure}

We use a convolutional word embedding model to classify the tweets into these 22 different topics and to predict their sentiment (positive, negative or neutral). From the perspective of the model, topic and sentiment are the same things since they are both labels for tweets. The convolutional embedding model (see Figure \ref{topic_cnn}) assigns a $d$ dimensional
vector to each of the $n$ words of an input tweet resulting in a matrix of size $n \times d$. Each of these vectors are initialized with uniformly distributed random numbers i.e. $x_i \in {R}^d$. The model, though randomly initialized, will eventually learn a look-up matrix ${R}^{|V|\times d}$ where $|V|$ is the vocabulary size, which represents the word embedding for the words in the vocabulary. 

A convolution layer is then applied to the $n \times d$ input tweet matrix, which takes into consideration all the successive windows of size $l$, sliding over the entire tweet. A filter $w \in {R}^{h\times d}$ operates on the tweet to give a feature map $c\in {R}^{n-l+1}$. We apply a max-pooling function \cite{collobert2011natural} of size $p=(n-l+1)$ shrinking the size of the resultant matrix by $p$. In this model, we do not have several hierarchical convolutional layers - instead we apply convolution and max-pooling operations with $f$ filters on the input tweet matrix for different window sizes ($l$).

The vector representations derived from various window sizes can be interpreted as prominent and salient n-gram word features for the tweets. These features are concatenated to create a vector of size $f \times L$, where $L$ is the number of different $l$ values , which is further compressed to a size $k$, before passing it to a fully connected softmax layer. The output of the softmax layer is the probability distribution over topic/sentiment labels. Table \ref{qe} shows a few examples of top terms being associated with various topics and sentiment categories from the output of the softmax layer. Next, two dropout layers are used, one on the feature concatenation layer and other on the penultimate layer for regularization ($\rho=0.5$). Though the same model is employed for both topic and sentiment classification tasks, they have different hyperparameters as shown in Table \ref{topic_sent}.

\begin{table}[h]
\centering
\small

\begin{tabular}{|l|l|}
\hline %
\emph{Topic/Sent} & \emph{Top Terms} \\ \hline 
Health care & medicaid, obamacare, pro obamacare, \\ & medicare, anti obamacare, repealandreplace \\\hline
Racial Issues & blacklivesmatter,civil rights, racistremarks,\\&  quotas, anti blacklivesmatter, tea party racist \\\hline
Guns & gunssavelives, lapierre, pro gun, gun rights, \\ & nra, pro 2nd amendment, anti 2nd amendment \\\hline
Immigration & securetheborder, noamnesty, antiimmigrant, \\& norefugees, immigrationreform, deportillegals \\\hline
Jobs & minimum wage, jobs freedom prosperity, \\ & anti labor, familyleave, pay equity \\\hline
Positive & smiling, character, courage, blessed, safely,\\ & pleasure, ambitious, approval, recommend \\\hline

\end{tabular}
\caption{Examples of top terms from vocabulary associated with a subset of the topics based on their probability distribution over topics and sentiments from the softmax layer.}
\label{qe} %
\end{table}

\begin{table}[]%
\centering
\small
\begin{tabular}{ |c|c|c|c|c| }
\hline %
\emph{Classifier} & \emph{Word}    & \emph{Penultimate Layer} \\
&   \emph{Embedding} \emph{(d)} & \emph{Size} \emph{(k)} \\\hline %
Topic & $300$ &  256  \\\hline
Sentiment &$200$ &  128 \\\hline

\end{tabular}
\caption{Hyperparameters based on cross-validation for topic and sentiment classifiers ($L=3$ i.e. $l\in \{2,3,4\}$, $f=200$ for both). }
\label{topic_sent} %
\end{table}

To learn the parameters of the model, as the training objective we minimize the cross-entropy loss. This is given by:
\begin{equation}\label{cat_ce} CrossEnt(p,q)=-\sum p(x)\log(q(x))\end{equation}
where p is the true distribution and q is the output of the softmax. This, in turn, corresponds to computing the negative log-probability of the true class. We resort to Adam optimization algorithm \cite{kingma2014adam} here as well.

Distance supervision was used to collect the training dataset for the model. The same annotator that identified the 22 election-related topics also created a list of ``high precision'' terms and hashtags for each of the topics. These terms were expanded using the same technique as was used for the ingest engine. The expanded terms were used to collect a large number of example tweets (tens of thousands) for each of the 22 topic. Emoticons and adjectives (such as happy, sad, etc) were used to extract training data for the sentiment classifier. As mentioned earlier about the election classifier, though distance supervision is noisy, the sheer number of training examples make the benefits outweigh the costs associate with the noise, especially when using the data for training deep neural networks.

We evaluated the topic and sentiment convolutional models on a set of 1,000 election-related tweets which were manually annotated. The topic classifier had an average (averaged across all classes) precision and recall of 0.91 and 0.89 respectively, with a weighted F-score of 0.90. The sentiment classifier had an average precision and recall of 0.89 and 0.90 respectively, with a weighted F-score of 0.89.


\section{Conclusions}
In this paper, we presented a system for detection and categorization of election-related tweets. The system utilizes recent advances in natural language processing and deep neural networks in order to--on a continuing basis--ingest, process and analyse all English-language election-related data from Twitter. The system uses a character-level CNN model to identify election-related tweets with high precision and accuracy (F-score of .92). These tweets are then classified into 22 topics using a word-level CNN model (this model has a F-score of .90). The system is automatically updated on a weekly basis in order to adapt to the new terms that inevitably enter the conversation around the election. Though the system was trained and tested on tweets related to the 2016 US presidential election, it can easily be used for any other long-term event, be it political or not. In the future, using our rumor detection system \cite{vosoughi2015automatic} we would like to track political rumors as they form and spread on Twitter.


\small
\bibliographystyle{aaai}
\bibliography{bib}
\end{document}